\documentclass[11pt]{article}

\usepackage{ACL2023}

\usepackage{times}
\usepackage{latexsym}

\usepackage[T1]{fontenc}

\usepackage[utf8]{inputenc}

\usepackage{microtype}
\usepackage{ifthen}
\usepackage{inconsolata}
\usepackage{algorithm}
\usepackage{algpseudocode}
\PassOptionsToPackage{table,dvipsnames}{xcolor}
\usepackage{xcolor}
\usepackage{colortbl}
\usepackage{booktabs} 
\usepackage{siunitx} 
\usepackage{amsmath} 

\usepackage{amsmath}
\usepackage{graphicx}
\usepackage{multirow}
\usepackage{multicol}
\usepackage{rotating}
\usepackage[caption=false,font=footnotesize]{subfig}
\usepackage{fancyhdr}
\usepackage{lipsum} 

\title{Progressively Modality Freezing for Multi-Modal Entity Alignment}

\author{
Yani Huang$^{1}$, Xuefeng Zhang$^{1}$, Richong Zhang$^{1,2}$\thanks{\ \ Corresponding author.}, Junfan Chen$^{3}$, Jaein Kim$^{1}$ \\
$^{1}$CCSE, School of Computer Science and Engineering, Beihang University, Beijing, China \\
$^{2}$Zhongguancun Laboratory, Beijing, China \\
$^{3}$School of Software, Beihang University, Beijing, China \\
\texttt{\{huangyn, zhangxf, jaein\}@buaa.edu.cn} \\
\texttt{\{zhangrc, chenjf\}@act.buaa.edu.cn}
}

\begin{document}
\maketitle
\begin{abstract}
Multi-Modal Entity Alignment aims to discover identical entities across heterogeneous knowledge graphs. 
While recent studies have delved into fusion paradigms to represent entities holistically, the elimination of features irrelevant to alignment and modal inconsistencies is overlooked, which are caused by inherent differences in multi-modal features.
To address these challenges, we propose a novel strategy of progressive modality freezing, called PMF, that focuses on alignment-relevant features and enhances multi-modal feature fusion. 
Notably, our approach introduces a pioneering cross-modal association loss to foster modal consistency.
Empirical evaluations across nine datasets confirm PMF's superiority, demonstrating state-of-the-art performance and the rationale for freezing modalities. Our code is available at~\href{https://github.com/ninibymilk/PMF-MMEA}{https://github.com/ninibymilk/PMF-MMEA}.
\end{abstract}

\section{Introduction}\label{sec:intro}
Multi-modal Knowledge Graphs (MMKGs) integrate various modalities, including text, vision, and structural data, to provide 
comprehensive insights into knowledge. This integration underpins a range of applications,  from question answering~\cite{zhu2015building} and information retrieval~\cite{dietz2018utilizing,yang2020biomedical}, to recommendation systems~\cite{sun2020multi}. Multi-Modal Entity Alignment (MMEA) aims to identify identical entities across heterogeneous MMKGs, which is essential for the integrity of the knowledge represented within these KGs.
\begin{figure}
    \centering
    \includegraphics[width=1.0\linewidth]{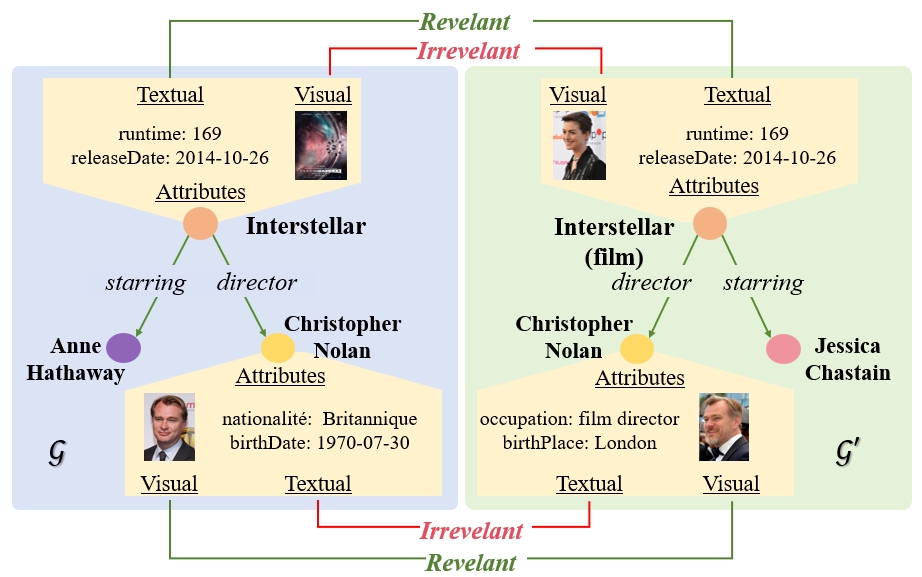}
    \caption{
    Illustration of irrelevant vs. relevant features in multi-modal knowledge graphs.
    }
    \label{fig:intro}
\end{figure}
The essence of MMEA lies in identifying feature commonalities across entities from varied modalities to determine their alignment. 
The diversity in KG construction introduces potential mismatches in multi-modal features of entities meant to be aligned. For example, Figure~\ref{fig:intro} depicts how the entity ``Interstellar (film)'' might be represented in one knowledge graph \(\mathcal{G} \)  by a poster image of a spaceship, whereas in another knowledge graph \(\mathcal{G'} \), by a portrait of Anne Hathaway, the starring actress. Although both images are related to ``Interstellar (film)'', their disparate perspectives can weaken the semantic relationship, challenging the alignment task. 
This scenario underscores the problem of {\it alignment-irrelevant features}, distinctly different features that complicate accurate entity alignment. While recent studies~\cite{lin2022multi, chen2023meaformer} have employed attention mechanisms to calculate cross-modal weights at both the KG and entity levels, they overlook the critical step of evaluating modality relevance, thereby neglecting to exclude these irrelevant features.

Another obstacle in MMEA involves ensuring semantic consistency across different modalities. Achieving consistent representations across modalities is crucial for effective entity alignment.
Existing methods~\cite{lin2022multi,chen2023meaformer,chen2023rethinking} utilize contrastive learning to minimize the intra-modal inconsistencies among aligned entities, yet overlook the {\it cross-modal inconsistencies}. 
Furthermore, the presence of alignment-irrelevant features exacerbates these inconsistencies, complicating the task of achieving consistent representations across modalities.

To address the aforementioned challenges, we propose 
a novel method for multi-modal entity alignment, named Progressive Modality Freezing (PMF), designed to effectively filter out irrelevant features. PMF is structured around three key parts: multi-modal entity encoders, progressive multi-modality feature integration, and a unified training objective for MMEA. 
Initially, PMF separately encodes entity features from each modality, allowing for flexible adjustment of input modalities.
Following this, the method employs a progressive approach to selectively freeze features deemed less relevant for alignment, simultaneously integrating valuable multi-modal features based on their alignment relevance.
The culmination of our strategy is a unified training objective, aimed at minimizing the discrepancies both within individual KGs and across different modalities.

To encapsulate, our research contributes in the following three-folds:
\begin{itemize}
    \item We propose the Progressive Modality Freezing (PMF) strategy to tackle the challenge of irrelevant features in MMEA. By assigning alignment-relevance scores to modalities, our method progressively freezes features that are alignment-irrelevant, while seamlessly integrating beneficial multi-modal information, ensuring the emphasis remains on features of utmost relevance.
    \item We are at the forefront of employing contrastive learning to address modality consistency across multiple modalities, accompanied by a unified training objective to enhance cross-modal consistency.
    \item We confirm the effectiveness of PMF across diverse datasets and experimental settings, where it demonstrates superior performance, achieving state-of-the-art in the field. Our thorough analysis further elucidates the advantage of the feature-freezing strategy.
\end{itemize}

\section{Related Work}

\subsection{Multi-modal Knowledge Graph Representation Learning}
The field of MMKG representation learning forms a critical foundation for tasks such as entity alignment and link prediction, transitioning from traditional single-modal knowledge graph methods to more complex that leverage diverse data types.
Early efforts expanded upon established KG representation learning methods for single-modal knowledge graphs, like those introduced by ~\citet{xie2016image} and ~\citet{mousselly2018multimodal}, adapting translational KG embedding techniques ~\cite{bordes2013translating} to incorporate multi-modal data effectively. 
Subsequent research has delved deeper into the impact of multi-modal information in representation learning. 
~\citet{pezeshkpour2018embedding} implemented an adversarial method to tackle the challenge of missing data within MMKGs, while ~\citet{fang2022contrastive} explored a contrastive learning approach to leverage the rich, varied data and heterogeneous structures in MMKGs. These advancements highlight the field's move towards more sophisticated, data-inclusive methodologies.

\subsection{Multi-Modal Entity Alignment}
The exploration of MMEA began with MMKG ~\cite{liu2019mmkg}, which provided a dataset facilitating entity embedding through linked images. 
Building on this, MMEA~\cite{chen2020mmea} advanced the field by fusing relational triples, images, and numerical attributes to generate entity embeddings.

Given the distinctiveness of images as a modality, utilizing visual data in alignment processes has emerged as a key focus in MMEA. For instance, EVA~\cite{liu2021visual} exploited visual similarities for unsupervised entity alignment, while MSNEA~\cite{chen2022multi} and UMAEA~\cite{chen2023rethinking} enhanced relational feature learning using visual cues to tackle the challenges posed by missing and ambiguous visual data.

Central to MMEA is the integration of multi-modal information. MEAformer~ \cite{chen2023meaformer} employed a dynamic weighting transformer for integration, whereas MCLEA ~\cite{lin2022multi} aimed at bridging the modality gap via contrastive learning. ACK-MMEA ~\cite{li2023attribute} introduced a strategy for normalizing multi-modal attributes, ensuring consistency across modalities. 

Iterative learning has emerged as a powerful strategy for refining entity alignment with methods like XGEA~\cite{xu2023cross} and PSNEA~\cite{ni2023psnea} employing pseudo-labeling to mitigate scarcity of initial seeds. 

Despite these advancements, existing methods neglect the necessity of filtering out alignment-irrelevant features and addressing cross-modal inconsistencies. 
Our method addresses these gaps by selectively freezing less pertinent modalities within an entity to enhance the alignment process.

\section{Model}
\begin{figure*}[!t]
    \centering
    \includegraphics[width=2.1\columnwidth]{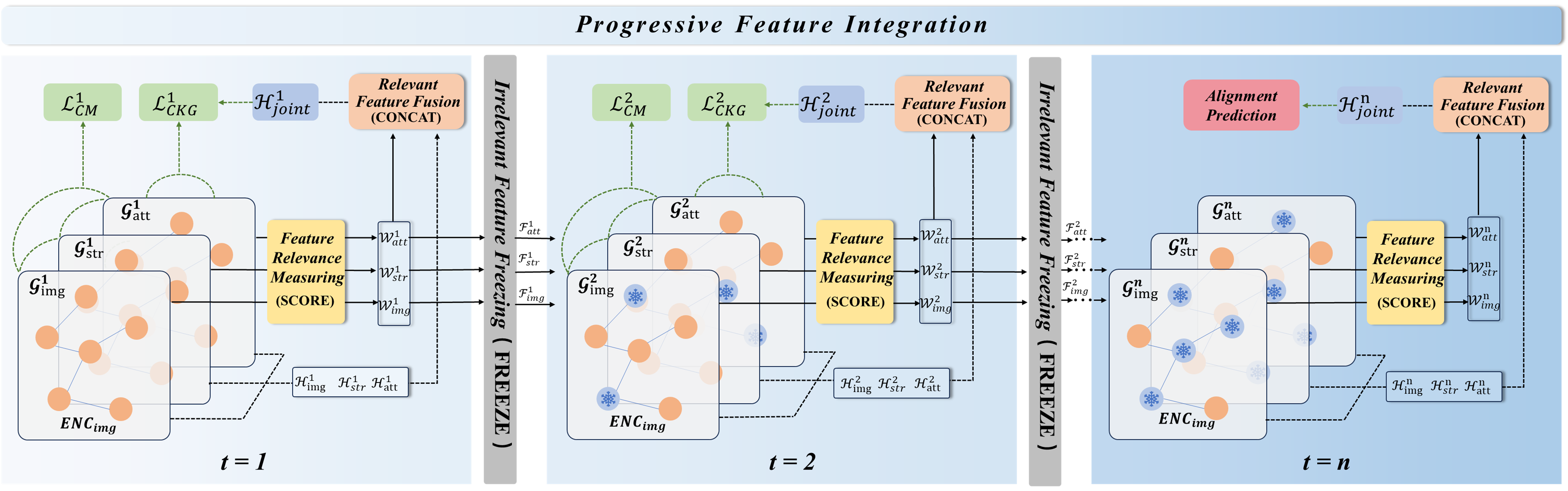} 
    \caption{Overview of the PMF Model. The framework consists of three components: the Multi-Modal Entity Encoder (\&\ref{Multi-Modal Entity Encoder}), Progressive Multi-Modality Feature Integration (\&\ref{Progressive Multi-Modality Feature Integration}), and Cross-Graph Contrastive Learning (\&\ref{Multi-Modal Entity Alignment Objective}). At epoch t, the multi-modal encoder {\textit{ENC$_m$}} transforms raw inputs from each modality graph into entity embeddings {\textit{$\mathcal{H}^t_m$}}. {\textit{Progressive Feature Integration}} (\&\ref{Progressive Features Integration}) occurs during training, employing {\textit{Irrelevant Feature Freezing}} (\&\ref{Irrelevant Feature Freezing}) and performing {\textit{Relevant Feature Fusion}} (\&\ref{Relevant Features Fusion}) guided by {\textit{Relevant Feature Measuring}} (\&\ref{Feature Relevance Measuring}). The model is optimized using Cross-KG Alignment Loss {\textit{$\mathcal{L}^t_{CKG}$}} (\&\ref{Cross-KG Alignment Loss}) and Cross-Modality Association Loss {\textit{$\mathcal{L}^t_{CM}$}} (\&\ref{Cross-Modality Association Loss}).}
    \label{fig:overview}
\end{figure*}
We propose a multi-modal entity alignment framework based on progressively freezing entities with irrelevant features and contrastive learning. As shown in Figure~\ref{fig:overview}, the framework consists of three parts: multi-modal entity encoder, progressive multi-modality feature integration, and cross-graph contrastive learning. First, a multi-modal encoder is used to transform the raw input of each modality graph into entity embeddings. Then, we integrate modality features progressively during training, which gradually freezes alignment-irrelevant features and fuse the modality graphs. Finally, the model is optimized through contrastive loss between different knowledge graphs and different modality graphs.

\subsection{Preliminaries} \label{preliminaries}

\paragraph{MMKG Definition} A multi-modal knowledge graph (MMKG) is defined as $\mathcal{G}=\{\mathcal{E},\mathcal{R},\mathcal{T},\mathcal{A}\}$, where $\mathcal{E}$ is a set of entities, $\mathcal{R}$ is a set of relations between entities, $\mathcal{T}\subseteq \mathcal{E}\times\mathcal{R}\times\mathcal{E}$ denotes a set of triples in the knowledge graph, and $\mathcal{A}$ is a set of multi-modal attributes of entities (e.g., textual descriptions, images, etc.).
For modality $m\in {\cal M}$, where $\cal M$ is the set of modalities, we define a modality graph $\mathcal{G}_{m}$ that captures a unique aspect pertinent to the knowledge graph. More specifically, $\mathcal{G}_{m}=\{\mathcal{E},\mathcal{R},\mathcal{T},\mathcal{A}_m\}$, where $\mathcal{A}_m$ is the set of attributes associated with modality $m$. 

\paragraph{Task Description} 
In the context of multi-modal entity alignment (MMEA), consider two MMKGs $\mathcal{G}=\{\mathcal{E},\mathcal{R},\mathcal{T},\mathcal{A}\}$ and $\mathcal{G'}=\{\mathcal{E'},\mathcal{R'},\mathcal{T'},\mathcal{A'}\}$.
The set $\mathcal{S}= \{(e_i,e_j)|e_i\in\mathcal{E},e_j\in\mathcal{E'}\}$ denotes pre-established pairs of entities known to correspond to the same real-world object across both graphs. 
In a supervised learning setting, a model is trained with part of $\cal S$ and then tasked to predict the alignments of the remaining entity pairs.

\subsection{Multi-Modal Entity Encoder}
\label{Multi-Modal Entity Encoder}

In the Multi-modal Entity Encoder module, we process the features of entities from each KG according to their respective modalities. Given an entity $e_i \in \mathcal{G}_m$, its feature is denoted by $e_i^m$. The encoding process for each modality of an entity is defined by the following equation:
\begin{equation}
    {h_i^m} = {\rm ENC}_m(\Theta_m, e_i^m),
\end{equation}
where ${\rm ENC}_m$ represents the encoder designed for modality $m$ with its associated learnable parameter $\Theta_m$.
While our framework allows the integration of any modality,  for a fair comparison, we incorporate four modalities: structure ({\it str}), relation ({\it rel}), attribute ({\it att}), and image ({\it img}). Details of ${\rm ENC}_m$ are attached in Appendix~\ref{app:Encoder Details}

\subsection{Progressive Multi-Modality Feature Integration}
\label{Progressive Multi-Modality Feature Integration} 
The process of integrating multi-modal features in a stepwise manner unfolds in three distinct phases. Initially, the process involves calculating an alignment-relevant score for each modality graph (e.g., $\mathcal{G}_m$) to discern features that are irrelevant for alignment from those that are crucial. Following this, certain features are selectively `frozen' based on modality scores to mitigate the inclusion of extraneous features. Finally, features from various modality graphs are fused to align multi-modal entities. This approach is executed iteratively and progressively to identify and incorporate beneficial features into the alignment process throughout the training phase, ensuring a refined alignment outcome.

\subsubsection{Feature Relevance Measuring} 
\label{Feature Relevance Measuring}
We assess the likelihood that features from a given modality of an entity will contribute to the alignment process, thereby enabling the identification of potential alignment-irrelevant features within each modality graph. The alignment-relevant scores for a modality graph $\mathcal{G}_m$ are denoted as $\mathcal{W}_m$, calculated by the following equation:
\begin{eqnarray}
     \mathcal{W}_m = {\rm SCORE}(\mathcal{G}_m,\mathcal{G}_m'),
\end{eqnarray}
where ${\rm SCORE}(\cdot)$ represents a score function. Specifically, for an entity $e_i^m$ from $\mathcal{G}_m$ and considering a corresponding modality graph $\mathcal{G}_m'$ from target knowledge graph $\mathcal{G}'$, the alignment-relevant score is computed as:
\begin{eqnarray}
\label{exp:confidence}
    w_i^m &=& {\rm Relu}\left(\frac{\alpha_i^m - \delta}{\max\limits_{e_j \in \mathcal{E}}(\alpha_j^m) - \delta}\right), \\
    \alpha_i^m &=& \max\limits_{e_k \in \mathcal{E'}}\left(\frac{h_i^m \cdot h_k^m}{|h_i^m||h_k^m|}\right),
\end{eqnarray}
where $\delta$ is a dynamic threshold. The collection of $w_i^m$ for all entities constitutes $\mathcal{W}_m$. In a parallel manner, the alignment-relevant scores $\mathcal{W}_m'$ for $\mathcal{G}_m'$ are computed similarly.

The underlying rationale for the alignment-relevant score is to ascertain whether an entity from the current modality graph $\mathcal{G}_m$ exhibits similarity with entities from the modality graph $\mathcal{G}_m'$. Should the features $h_i^m$ of an entity $e_i$ in $\mathcal{G}_m$ lack similar counterparts in $\mathcal{G}_m'$, the correlation between $h_i^m$ and the entity aligned with $e_i$ in $\mathcal{G}_m'$ would consequently be weak. As such, $h_i^m$ is less likely to contribute positively to the alignment process. To facilitate a reasonable measurement of feature similarity within a modality, we normalize this metric using the highest similarity across all entities from the same modality graph. This alignment-relevant score is subsequently utilized to guide the processes of modality feature freezing and fusion.

\subsubsection{Irrelevant Feature Freezing}
\label{Irrelevant Feature Freezing}
To minimize the detrimental impact of alignment-irrelevant features on the training process, we introduce a feature-freezing strategy. This approach selectively prevents back-propagation for specific entities within particular modality graphs, thereby preserving the integrity of the training regime. 
Formally, we define $\mathcal{G}_{m}^t$ by extending $\mathcal{G}_{m}$ with a function ${\cal F}_m^t: \mathcal{E}\rightarrow \{0, 1\}$, to characterize whether entity $e$ is frozen at step $t$. 
More specifically,  $\mathcal{G}_{m}^t=\{\mathcal{E},\mathcal{R},\mathcal{T},\mathcal{A}_m, {\cal F}_m^t\}$, where for each entity $e$ in $\mathcal{E}$,  $e$ is frozen 
only if ${\cal F}_m^t(e)= 0 $. Correspondingly, we refer ${\cal H}_m^t$ and ${\cal W}_m^t$ to the representation of all entities and their alignment-relevant score at step $t$, respectively. All elements in vector ${\cal F}_m^0$ are initialized as $1$.

The formalization of the entity freezing process is given by: 
\begin{equation}
    {\mathcal{F}}_m^{t}= {\rm FREEZE}(\mathcal{W}_m^t,\mathcal{G}^{t}_m).
\end{equation}
At step $t$, when the alignment-relevant score $w_i^m$ for entity $e_i$ from $\mathcal{G}_m^t$ is zero, it is inferred that this entity does not contribute positively towards the training of the alignment model. Consequently, the gradient update for features $h_i^m$ associated with such an entity is halted during back-propagation. The freezing operation for an entity $e_i^m$, contingent upon its score $w_i^m$, is defined as:
\begin{equation}
{\rm FREEZE}(e_i^m) =
\begin{cases}
0 & \text{if } w_i^m = 0, \\
1 & \text{otherwise.}
\end{cases},
\end{equation}
where $0$ indicates "stop gradient", ensuring that the gradient flow is interrupted for entities deemed irrelevant, with $w_i^m\in \mathcal{W}_m^t$. ${\rm FREEZE}(\cdot)$ systematically applies across all entities within $\mathcal{G}_m^t$. This feature freezing is enacted across all modality graphs of each KG (e.g., $\mathcal{G}$ and $\mathcal{G}'$) to optimize the alignment model's focus on beneficial features.


\subsubsection{Relevant Feature Fusion}
\label{Relevant Features Fusion}
Recognizing that the significance of the features of each modality may vary across entities, we leverage the alignment-relevant scores introduced in Eq.~\ref{exp:confidence} as a guide during the fusion process.
\begin{equation}
\mathcal{H}_{joint}^t = \mathop{\rm CONCAT}\limits_{m \in {\cal M}}({\cal W}_m^t,\mathcal{G}_m^t), 
\end{equation}
where $\rm CONCAT(\cdot)$ symbolizes the fusion operator acting on all modality graphs $\mathcal{G}_m^t$, integrating them into a joint representation $\mathcal{H}_{joint}^t$.

In practice, for each entity $e_i$, the representations $h_i^m$ are weighted by their respective confidence $w_i^m$, ensuring that modalities with higher relevance scores have a greater influence in the fused representation. Accordingly, at step $t$, this fusion process is defined as follows: 
\begin{equation}
h^{joint}_i = \mathop{\mathrm{\oplus}}\limits_{m \in \mathcal{M}}([w_i^m\cdot h_i^m]), 
\end{equation}
where $\mathrm{\oplus}$ denotes the vector concatenation operation, and $h^{joint}_i$ represents the composite feature embedding for entity $e_i$, synthesized from all modalities. The collective $h^{joint}_i$ across all entities at step $t$ forms $\mathcal{H}_{joint}^t$.

\subsubsection{Progressive Feature Integration}
\label{Progressive Features Integration}


By integrating information from each modality graph, we can progressively optimize entity alignment. This is done by iteratively refining feature relevance, and gradually isolating alignment-irrelevant features from overt to subtle ones. Initially, basic representations are used to gauge modality confidence and selectively freeze modality features, employing a lower threshold to exclude overtly irrelevant features due to nascent modality representations. As training advances, the model accumulates a deeper understanding of modal information, systematically increasing the threshold to prune less obvious features independent of alignment goals.


The process unfolds as follows: Assuming the model has been trained for $t$ epochs, in epoch $t+1$, we compute the alignment-relevant scores $\mathcal{W}_m^{t+1}$ using the current stage of $\mathcal{G}_m^t$ and $\mathcal{G}_m^{'t}$. Concurrently, the threshold $\delta$ for discerning feature relevance is increased at a predefined rate. Based on the newly calculated scores $\mathcal{W}_m^{t+1}$, entities are frozen to reflect the updated insights, updating each modality graph to $\mathcal{G}_m^{t+1}$. Subsequently, these refined modality graphs are merged, leveraging the latest alignment-relevant scores to produce the joint graph representation $\mathcal{H}_{joint}^{t+1}$, as formalized in Equation (2), (5) and (7).

Comparatively, alternatives like ``early integration'' or ``late integration'' present limitations. Early integration might only freeze superficially irrelevant features due to the premature stage of training, which can be challenging to correct in later stages. Our progressive strategy synergistically blends the benefits of both approaches, ensuring a balanced and effective integration of modality information throughout the training lifecycle. 
 
\subsection{Multi-Modal Entity Alignment Objective} 
\label{Multi-Modal Entity Alignment Objective}
Our framework adopts contrastive learning as a pivotal strategy to extract and leverage alignment information across knowledge graphs and to discern semantic associations between various modality graphs. To accomplish this, we have developed a composite loss function that integrates both cross-KG alignment and cross-modality association within a unified learning objective. The overall loss function at step $t$ is defined as follows:

\begin{equation}
     \mathcal{L}^t = \mathcal{L}^t_{\rm CM} + \mathcal{L}_{\rm CKG}^t,
\end{equation}
where $\mathcal{L}_{\rm CKG}^t$ is the alignment loss across different KGs, and $\mathcal{L}_{\rm CM}^t$ denotes the association loss across modalities within each KG. 

\subsubsection{Cross-Modality Association Loss}
\label{Cross-Modality Association Loss}
One significant source of inconsistencies across modalities is alignment-irrelevant features. Freezing features within each modality graph allows our framework to more effectively draw together the remaining modal features that are beneficial for alignment.

Specifically for each entity $e_i$, we compute the contrastive loss between pairs of modality graphs $\mathcal{G}_p$ and $\mathcal{G}_q$. The loss function aims to minimize the distance between positive pairs $(e_i^p,e_i^q)$ and maximize the separation between negative pairs $(e_i^p,e_j^q)$ and $(e_j^p,e_i^q)$. Here $e_i$ and $e_j$ are entities in the same knowledge graph ${\cal G}$. For each positive pair $(e_i^p,e_i^q)$, the cross-modality loss at step $t$ is defined as: 
\begin{equation}
\small
    {l}^t_{\rm CM}(e_i^p, e_i^q) = \frac{\exp(h_i^{p}\cdot h_j^{q}/\tau)}{\exp(h_i^{p}\cdot h_j^{q}/\tau)+\!\!\!\!\!\!\sum\limits_{(e_j^p,e_k^q)\in{N_{i,p,q}^{-}}}\!\!\!\!\!\!\!\!\!\!\!\!\exp(h_i^{p}\cdot h_j^{q}/\tau)},
\end{equation}
where $h_i^{p}{\in}{\cal H}_p^t$, $h_j^{q}{\in}{\cal H}_q^t$,  and $\exp(h_i^{p}{\cdot}h_j^{q}/\tau)$ are scaled by a temperature factor $\tau$, and set $N_{i,p,q}^{-}$ represents negative samples, encompassing all other entity pairs from the modality graphs $\mathcal{G}_p$ and $\mathcal{G}_q$.
\looseness=-1

The framework computes the contrastive loss for entities across each modality graph of the two KGs. The final cross-modality loss is then determined by:
\begin{equation}
\label{cross-modal loss}
\small
    \mathcal{L}^t_{\rm CM} = \sum_{p,q\in {\cal M}} \sum_{e_i\in {\mathcal{E}\cup\mathcal{E'}}}  \beta_p \beta_q \left(-\log l^t_{\rm CM}(e_i^p, e_i^q)\right),
\end{equation}
where $\beta_p$ and $\beta_q$ are hyper-parameters associated with modality $p$ and $q$, adjusting the learning rate for features of the corresponding modality.

\subsubsection{Cross-KG Alignment Loss}
\label{Cross-KG Alignment Loss}
Pre-aligned entity pairs serve as crucial supervision signals in multi-modal entity alignment. Our approach employs supervised contrastive learning to elucidate the alignment relations between entities from corresponding modality graphs of different KGs. We formalized this supervised contrastive learning process at step $t$ as follows:
\begin{equation}
\small
\begin{aligned}
    {l}^t_{ckg}(e_i^m, e_j^m) = \frac{\exp(h_i^m{\cdot}h_j^m/\tau)}{\exp(h_i^m{\cdot}h_j^m/\tau) + \!\!\!\sum\limits_{e_k\in{N_{i}^{-}}}\!\!\!\exp(h_i^m{\cdot}h_k^m/\tau)},
    \label{eq:mea-gamma}
\end{aligned}
\end{equation}
where $N_{i}^{-}$ encompasses a set of in-batch negative samples for $e_i^m$.

Given the undirected nature of entity alignment, we compute the contrastive loss considering both directions across KGs, which is defined as:

\begin{equation}
\begin{aligned}
    \mathcal{L}^t_{CKG} = \sum_{m \in \mathcal{M}^{+}} \sum_{(e_i, e_j) \in \mathcal{S}} 
    & -\frac{1}{2} \log \left( l^t_{ckg}(e_i^m, e_j^m) \right. \\
    & \left. + l^t_{ckg}(e_j^m, e_i^m) \right),
\end{aligned}
\end{equation}

where ${\cal M}^{+}$ is the set of modalities, including $\cal M$ and ``{\it joint}'' as a special modality that encapsulates the fusion features of multi-modal information. Note that the representations of modality ``{\it joint}'' is taken from ${\cal H}^t_{joint}$.

\subsection{Alignment Prediction}
\label{Alignment Prediction}
The alignment of entities is executed sequentially, focusing on the similarity between their fused multi-modal embeddings. Specifically, for each entity $e_i$ within the joint modality graph $\mathcal{G}_{joint}$ that remains unaligned, we seek its counterpart, $e_j$, in the corresponding joint modality graph $\mathcal{G}_{joint}'$. 
The sequence in which entities are aligned follows a greedy strategy.
The chosen $e_j$ is the unaligned entity that exhibits the highest cosine similarity score with $e_i$.

\section{Experiment}
\subsection{Experiment Setup}
\paragraph{Datasets}
To evaluate the effectiveness of our proposed method, we utilized three publicly available MMEA datasets.
\textbf{MMKG}~\cite{liu2019mmkg} features two subsets extracted from Freebase, YAGO, and DBpedia.  
\textbf{Multi-OpenEA}~\cite{sun2020benchmarking}, augmented with entity images from Google search queries, includes two multilingual subsets and two monolingual subsets.
\textbf{DBP15K}~\cite{sun2017cross,liu2021visual} comprises three subsets from DBpedia's multilingual version.
Seed alignments are designated for 20\% of MMKG and Multi-OpenEA entity pairs and 30\% for DBP15K entity pairs, aligning with proportions used in prior studies ~\cite{chen2020mmea,liu2021visual,lin2022multi,chen2023meaformer,chen2023rethinking}. 

\paragraph{Baselines}
We compare our approach against six leading multi-modal alignment methods: MMEA~\cite{chen2020mmea}, EVA~\cite{liu2021visual}, MSNEA~\cite{chen2022multi}, MCLEA~\cite{lin2022multi}, MEAformer~\cite{chen2023meaformer}, and UMAEA~\cite{chen2023rethinking}. 
We replicated MEAformer 
and UMAEA 
using their publicly available code to establish robust baselines. Although other notable methods like ACK-MMEA ~\cite{li2023attribute} exist, their exclusion was due to the lack of open-source code, ensuring fairness in our comparisons.

\paragraph{Evaluation Metrics}
We employ two metrics for evaluation. {Hits@1} (abbreviated as {H@1}) measures the accuracy of top-one predictions. {Mean Reciprocal Rank (MRR)} assesses the average inverse rank of the correct entity. Higher values of H@1 and MRR indicate better performance.

\paragraph{Implement Details}
The total training epochs are set to 250, with an option for an additional 500 epochs using an iterative training strategy. Our training regimen incorporates a cosine warm-up schedule (15\% step for LR warm-up), early stopping, and gradient accumulation, utilizing the AdamW optimizer 
with a consistent batch size of 3500.
Details of the experimental setup are available in the Appendix~\ref{sec:detailed setup}.\looseness=-1

\begin{table*}
\caption{Comparison of overall performance presenting both non-iterative and iterative results. The highest-performing baseline results are underlined, and any instances where our method sets a new state-of-the-art are highlighted in bold. Results with an asterisk (*) indicate our reproduction under identical settings.}
\centering
\resizebox{\textwidth}{!}{%
\begin{tabular}{cc|cc|cc|cc|cc|cc|cc|cc|cc|cc} 
\\ \hline
&\multirow{3}{*}{Model}& \multicolumn{4}{c|}{MMKG} & \multicolumn{8}{c|}{Multi-OpenEA} & \multicolumn{6}{c}{DBP15K} \\
&& \multicolumn{2}{c|}{FBDB15K} & \multicolumn{2}{c|}{FBYG15K} & \multicolumn{2}{c|}{EN-FR-V1} & \multicolumn{2}{c|}{EN-DE-V1} & \multicolumn{2}{c|}{D-W-V1} & \multicolumn{2}{c|}{D-W-V2} & \multicolumn{2}{c|}{ZH-EN} & \multicolumn{2}{c|}{JA-EN} & \multicolumn{2}{c}{FR-EN} \\

&& H@1 & MRR & H@1 & MRR & H@1 & MRR & H@1 & MRR & H@1 & MRR & H@1 & MRR & H@1 & MRR & H@1 & MRR & H@1 & MRR \\
\hline
\multirow{8}{*}{\rotatebox[origin=c]{90}{non-iterative}} & MMEA & .265 & .357 & .234 & .317 & - & - & - & - & - & - & - & - & - & - & - & - & - & - \\
& MSNEA & .114 & .175 & .103 & .153 & .692 & .734 & .753 & .804 & .800 & .826 & .838 & .873 & .609 & .685 & .541 & .620 & .557 & .643 \\
& EVA & .199 & .283 & .153 & .224 & .785 & .836 & .922 & .945 & .858 & .891 & .890 & .922 & .683 & .762 & .669 & .752 & .686 & .771 \\
& MCLEA & .295 & .393 & .254 & .332 & .819 & .864 & .939 & .957 & .881 & .908 & .928 & .949 & .726 & .796 & .719 & .789 & .719 & .792 \\
& MEAformer* & .418 & .519 & .327 & .418 & .836 & .882 & .954 & \underline{.971} & \underline{.909} & \underline{.933} & .944 & .962 & .771 & .835 & .764 & .834 & .772 & .841 \\
& UMAEA* & \underline{.454} & \underline{.552} & \underline{.355} & \underline{.451} & \underline{.847} & \underline{.891} & \underline{.955} & .970 & .905 & .930 & \underline{.948} & \underline{.967} & \underline{.800} & \underline{.860} & \underline{.801} & \underline{.862} & \underline{.818} & \underline{.877} \\
\rowcolor{gray!50} & 
\textbf{PMF} & \textbf{.539} & \textbf{.620} & \textbf{.459} & \textbf{.539} & \textbf{.912} & \textbf{.942} & \textbf{.973} & \textbf{.983} & \textbf{.955} & \textbf{.970} & \textbf{.981} & \textbf{.989} & \textbf{.835} & \textbf{.884} & \textbf{.835} & \textbf{.885} & \textbf{.850} & \textbf{.898} \\
\rowcolor{gray!30}&  
\textbf{improve↑} & \textbf{8.5\%} & \textbf{6.8\%} & \textbf{10.4\%} & \textbf{8.8\%} & \textbf{6.5\%} & \textbf{5.1\%} & \textbf{1.8\%} & \textbf{1.2\%} & \textbf{4.6\%} & \textbf{3.7\%} & \textbf{3.3\%} & \textbf{2.2\%} & \textbf{3.5\%} & \textbf{2.4\%} & \textbf{3.4\%} & \textbf{2.3\%} & \textbf{3.2\%} & \textbf{2.1\%} \\
\hline
\multirow{7}{*}{\rotatebox[origin=c]{90}{iterative}} & MSNEA & .149 & .232 & .138 & .210 & .699 & .742 & .788 & .835 & .809 & .836 & .862 & .894 & .648 & .728 & .557 & .643 & .583 & .672 \\
& EVA & .231 & .318 & .188 & .260 & .849 & .896 & .956 & .968 & .915 & .942 & .925 & .951 & .750 & .810 & .741 & .807 & .765 & .831 \\
& MCLEA & .395 & .487 & .322 & .400 & .888 & .924 & .969 & .979 & .944 & .963 & .969 & .982 & .811 & .865 & .805 & .863 & .808 & .867 \\
& MEAformer* & \underline{.578} & \underline{.661} & .444 & .529 & \underline{.903} & \underline{.935} & .963 & .977 & \underline{.954} & \underline{.970} & .969 & .981 & .847 & .892 & .842 & .892 & .845 & .894 \\
& UMAEA* & .561 & .660 & \underline{.463} & \underline{.560} & .895 & .931 & \underline{.974} & \underline{.984} & .945 & .965 & \underline{.973} & \underline{.984} & \underline{.856} & \underline{.900} & \underline{.857} & \underline{.904} & \underline{.873} & \underline{.917} \\

\rowcolor{gray!50} & 
\textbf{PMF} & \textbf{.624} & \textbf{.702} & \textbf{.543} & \textbf{.620} & \textbf{.923} & \textbf{.950} & \textbf{.980} & \textbf{.988} & \textbf{.960} & \textbf{.973} & \textbf{.986} & \textbf{.992} & \textbf{.867} & \textbf{.908} & \textbf{.866} & \textbf{.909} & \textbf{.879} & \textbf{.921} \\
\rowcolor{gray!30}&
\textbf{improve↑} & \textbf{4.6\%} & \textbf{4.1\%} & \textbf{8.1\%} & \textbf{6.0\%} & \textbf{2.0\%} & \textbf{1.5\%} & \textbf{0.6\%} & \textbf{0.4\%} & \textbf{0.6\%} & \textbf{0.3\%} & \textbf{1.3\%} & \textbf{0.8\%} & \textbf{1.1\%} & \textbf{0.8\%} & \textbf{0.9\%} & \textbf{0.5\%} & \textbf{0.6\%} & \textbf{0.4\%} \\ 
\hline
\end{tabular}
}
\label{tab:main_results}
\end{table*}

\subsection{Overall Results}

\begin{table}
\caption{Non-iterative model performance comparison utilizing surface information on DBP15K.}
\centering
\scalebox{0.7}{
\begin{tabular}{c|cc|cc|cc}    
\hline
\multirow{3}{*}{Model}&\multicolumn{6}{c}{DBP15K}\\
& \multicolumn{2}{c|}{ZH-EN} & \multicolumn{2}{c|}{JA-EN} & \multicolumn{2}{c}{FR-EN} \\ 
 & H@1 & MRR & H@1 & MRR & H@1 & MRR \\ \hline
MSNEA & .929 & .951 & .964 & .976 & .990 & .994 \\ 
EVA & .887 & .913 & .938 & .955 & .969 & .980 \\ 
MCLEA & .926 & .946 & .961 & .973 & .987 & .992 \\ 
MEAformer* & .944 & .962 & .976 & .985 & .990 & .994\\ 
UMAEA* & \underline{.945} & \underline{.963} & \underline{.978} & \underline{.986} & \underline{.990} & \underline{.994} \\ \hline
\rowcolor{gray!50} \textbf{PMF} & \textbf{.958} & \textbf{.973} & \textbf{.985} & \textbf{.990} & \textbf{.992} & \textbf{.995} \\ \hline
\end{tabular}
\label{tab:w/sf main results}
}
\end{table}

Table~\ref{tab:main_results} shows a comprehensive comparison between iterative and non-iterative methods across the three datasets, showcasing our model outperforms all nine sub-datasets.

Our approach significantly surpasses other non-iterative methods across various datasets. Specifically, on DBP15K, we see an improvement in H@1 of up to 3.5\% over the nearest competitor. The Multi-OpenEA datasets show enhancements in the range of 1.8\% to 6.5\%, while the FBDB15K and FBYG15K datasets experience substantial gains, with increases up to 10.4\%. These improvements highlight our method's capability to effectively filter out conflicting and lower-quality modality information, enhancing overall alignment accuracy.

To assess our model's performance with added textual data, we initialized text modality embeddings using surface representations from ~\cite{lin2022multi}. The incorporation of textual information markedly enhances entity alignment, as evidenced by the results in Table~\ref{tab:w/sf main results} for the DBP15K dataset. Despite all models benefiting from this supervisory signal, our method outperforms the baselines
and demonstrates its robustness in leveraging textual data for improved alignment. 

\subsection{Ablation Study }
\begin{figure}[ht]
    \centering
    \includegraphics[width=1\linewidth]{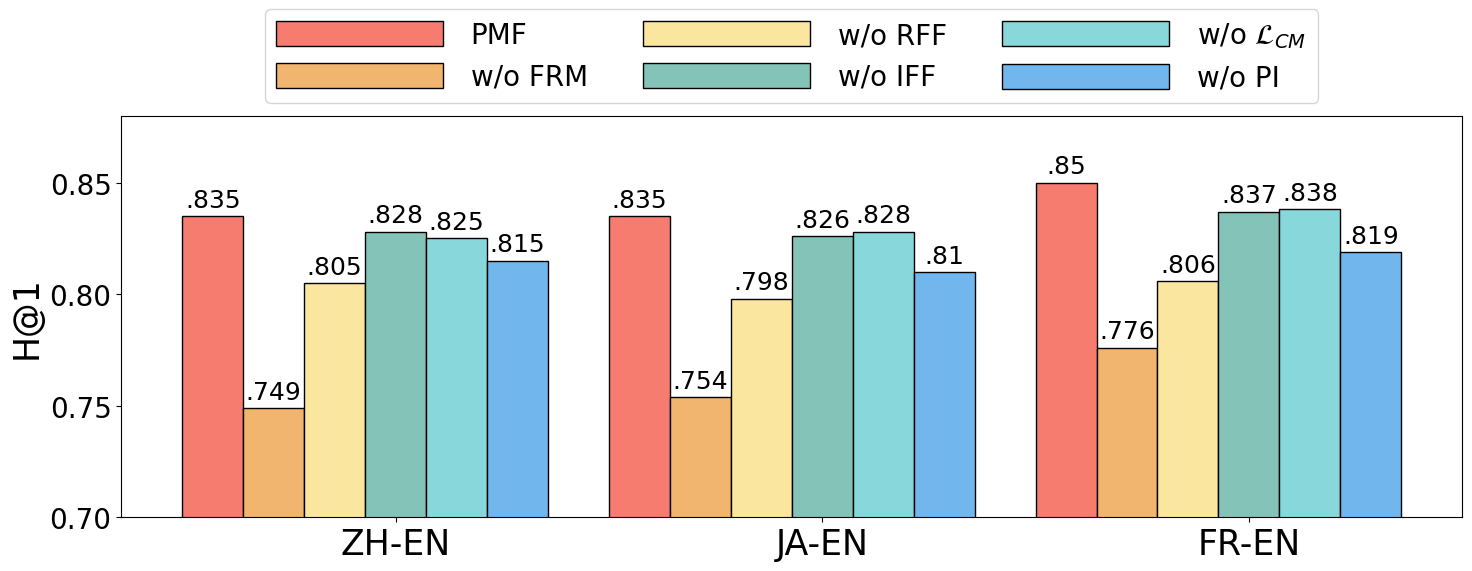} 
    \caption{Comparison of variants of PMF on DBP15K in terms of ${\rm H}@1$.}
    \label{fig:ablation_study_1}
\end{figure}

To assess the contribution of various components within our framework, we introduce five variants:
Feature Relevance Measuring (FRM), Irrelevant Feature Freezing (IFF), Relevant Feature Fusion (RFF), Progressive Integration (PI), and Cross-Modality  Association Loss (\(\mathcal{L}_{\rm CM}\)). By individually removing or modifying these components, their contributions to the performance were assessed. 
As depicted in Figure~\ref{fig:ablation_study_1}, the omission of the FRM component leads to a significant drop in H@1 performance across all datasets. For example, H@1 fell from 0.835 to 0.749 on DBP15K$_{\rm ZH-EN}$. The influence of FRM extends to both IFF and RFF components, with RFF showing a more significant impact on performance due to its integral role in modality fusion and its effect on the overall training loss.

Further, replacing our "progressive" integration strategy with a "static" one, where FRM occurs only once, resulted in a marked performance downturn. This underscores the significance of progressively identifying alignment-irrelevant features, a task challenging to accomplish in a single iteration.
Eliminating \(\mathcal{L}_{CM}\) also led to a decline in model performance. This underlines the effectiveness of cross-modality loss in fostering consistent entity representation learning across modalities.

\subsection{Modality Analysis}
\subsubsection{Impact of Different Modality }\label{sec:modality}
\begin{figure}[ht]
    \centering
    \includegraphics[width=1\linewidth]{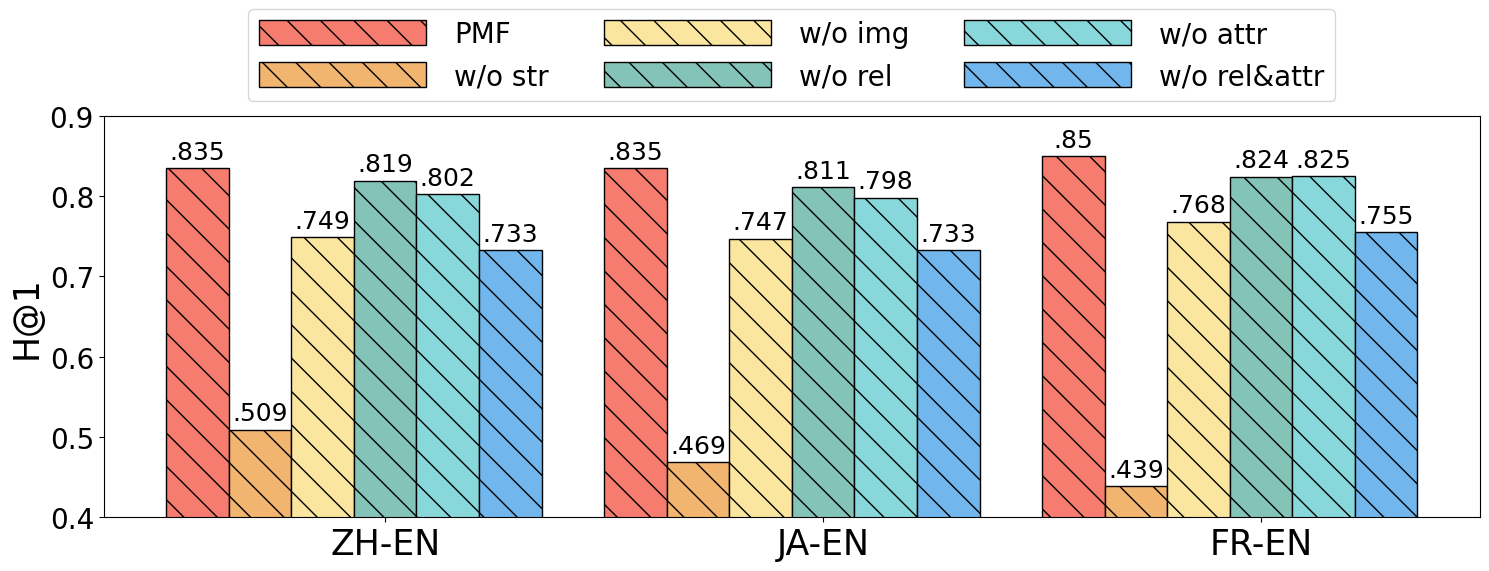}
    \caption{Impact of various modalities on DBP15K in terms of ${\rm H}@1$. }
    \label{fig:impact of modality}
\end{figure}

We examined the influence of modality information on entity alignment on DBP15K, as depicted in Figure~\ref{fig:impact of modality}. The absence of any modality notably hinders performance, with structure and image modalities showing the most significant effect. The foundational role of structures in KGs and the distinctive features provided by images underscore their importance in alignment.

A slight drop in performance is observed when either relation or attribute is excluded, attributed to the typically lower count of attributes and relations compared to entities, which may lead to attribute or relation overlaps among different entities. Conversely, a notable performance drop occurs when both are removed. This suggests a complementary relationship between attributes and relations that enhances alignment.

\subsubsection{Distribution of Modality Scores}
\begin{figure}[h]
    \centering
    \includegraphics[width=1\linewidth]{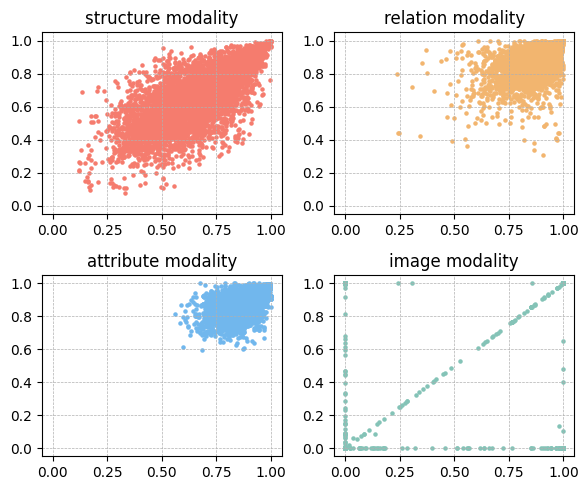}
    \caption{Distribution of relevance scores across modalities for pre-aligned entity pairs in ${\rm DBP15K}_{\rm FR-EN}$.}
    \label{fig:modality weight distribution}
\end{figure}
We examined the distribution of modality scores to understand the contribution of different modalities to entity alignment, as depicted in Figure~\ref{fig:modality weight distribution}. Each point in the figure represents an aligned entity pair, with axes reflecting their modality scores across source and target KGs. 

Attributes, relations, and structures show a widespread and increased distribution of scores, indicating their positive impact on alignment.
Conversely, image modality shows a distinct, concentrated pattern, highlighting the variability in image features between aligned entities due to the limited scope of images and inconsistencies in KG construction. This underscores the benefit of excluding less reliable image features. 
Moreover, the presence of data points near the axes suggests challenges with one-sided visual information.
Interestingly, a strong positive correlation emerges between modality scores of aligned entities, validating our modality scoring approach. This correlation implies that when a modality is less helpful for one entity in an aligned pair, it tends to be similarly low for the counterpart.


\subsection{Parameters Analysis}

\begin{figure}[h]
    \centering
    \includegraphics[width=1\linewidth]{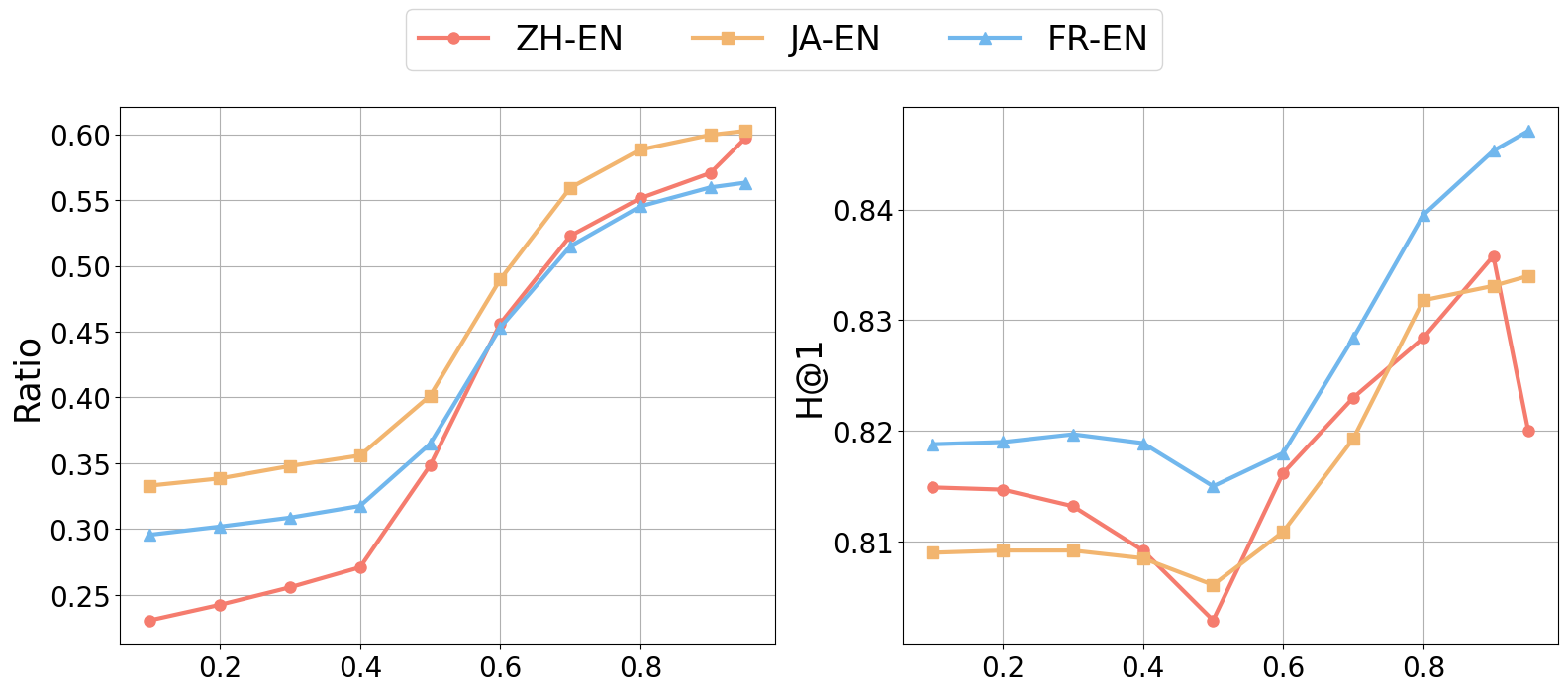}
    \caption{Impact of varying threshold $\delta$ on frozen ratio and H@1 performance: the x-axis represents $\delta$ values ranging from 0.1 to 0.95.}
    \label{fig:hyper-analy}
\end{figure}

To further investigate the impact of parameters on integration, we analyze how varying the threshold parameter $\delta$ impacts the frozen ratio in image modality and alignment performance on DBP15K. 
Figure~\ref{fig:hyper-analy} shows that 
when $\delta$ is below 0.4, minimal changes are observed in both the ratio and H@1 performance, likely due to the absence of images for certain entities. Beyond this point, 
as $\delta$ ascends, we note a gradual increase in the frozen ratio, paralleled by an improvement in H@1, despite the initial decline. 
This underscores the benefits of selectively and gradually freezing entities.
A notable drop in H@1 performance for some datasets occurs at $\delta=0.95$, indicating that images are beneficial for alignment and should not be frozen excessively.

Further experimental analysis results of the feature-freezing process of PMF, including additional experiments, are available in Appendix~\ref{app:detailed analysis}.
\subsection{Case Study}
\begin{table}[h]
\caption{Case study of representative frozen images on DBP15K\(_{\rm{FR-EN}}\) during training.}
\centering
\scalebox{0.7}{
\begin{tabular}{c|m{2cm}<{\centering}|m{2cm}<{\centering}|m{2cm}<{\centering}|m{2cm}<{\centering}} 
\hline
& \multicolumn{2}{c|}{Source-KG(French)} & \multicolumn{2}{c}{Target-KG(English)} \\ 
\cline{2-5}
 & Entity & Image & Entity & Image \\ 
\hline
\multirow{4}{*}{\rotatebox[origin=c]{90}{Early frozen}} & Louis XV &  \vspace{2pt}\includegraphics[width=0.1\textwidth]{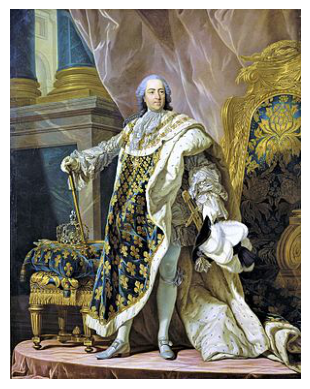} & Louis XV of France &  \includegraphics[width=0.1\textwidth]{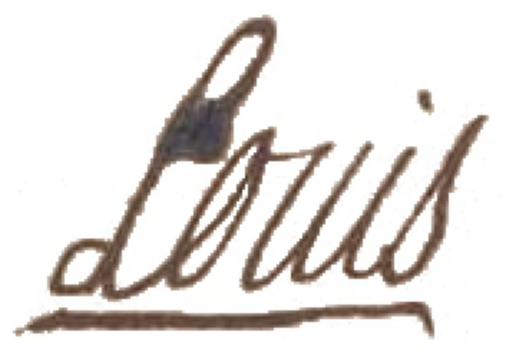}\\ 
\cline{2-5}
 & Président du Mexique  & \vspace{2pt}\includegraphics[width=0.1\textwidth]{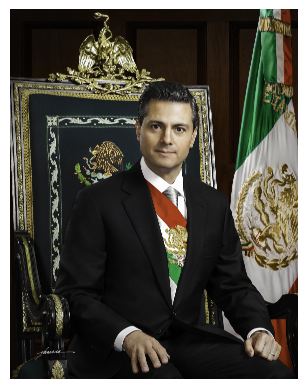} & President of Mexico & \includegraphics[width=0.1\textwidth]{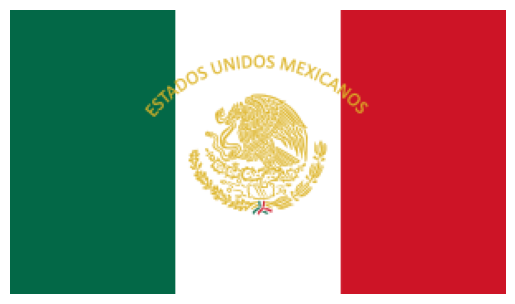}\\
\cline{2-5}
 & Université Stanford & \vspace{2pt}\includegraphics[width=0.1\textwidth]{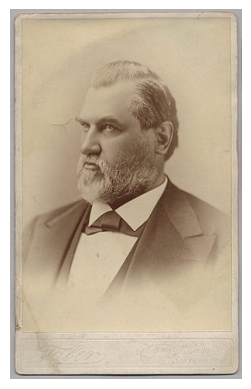} & Stanford University & \includegraphics[width=0.1\textwidth]{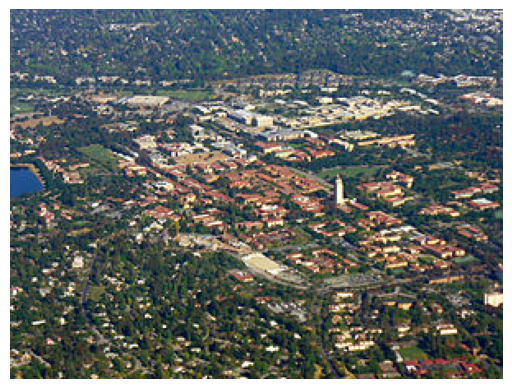}\\
\cline{2-5}
 & Disque compact & \includegraphics[width=0.1\textwidth]{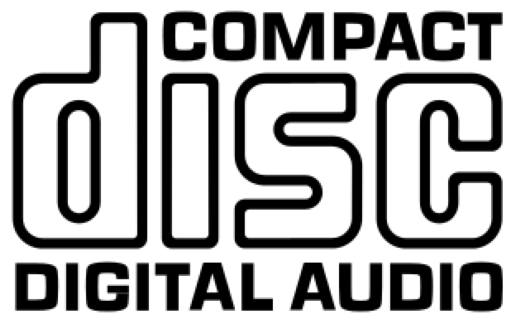} & Compact disc & \vspace{2pt}\includegraphics[width=0.1\textwidth]{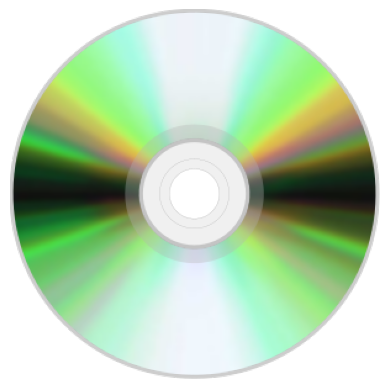}\\ 
\hline
\multirow{4}{*}{\rotatebox[origin=c]{90}{Late frozen}} & Juliana of the Netherlands &  \vspace{3pt}\includegraphics[width=0.1\textwidth]{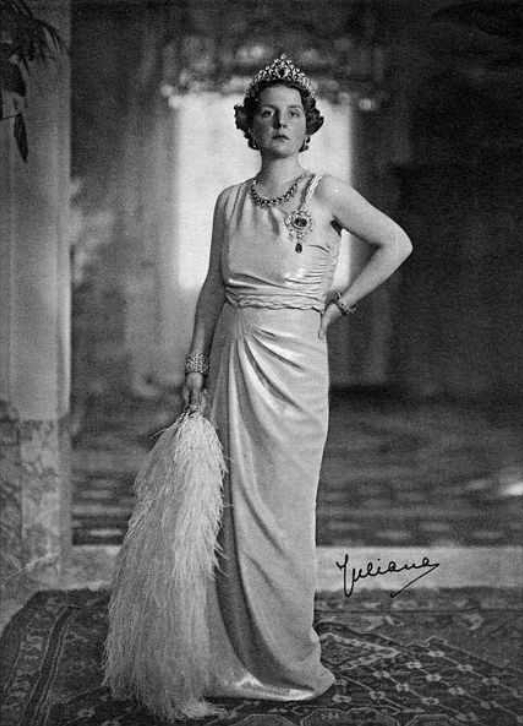} & Juliana of the Netherlands &  \includegraphics[width=0.1\textwidth]{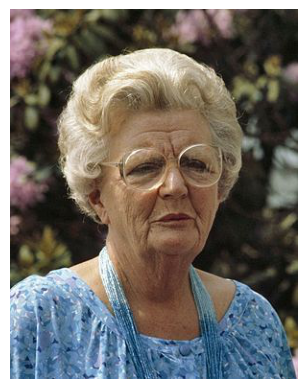}\\ 
\cline{2-5}
 & Brampton (Ontario)  &  \vspace{2pt}\includegraphics[width=0.1\textwidth]{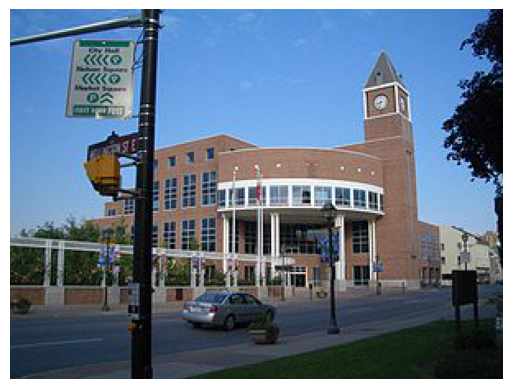} & Brampton &  \includegraphics[width=0.1\textwidth]{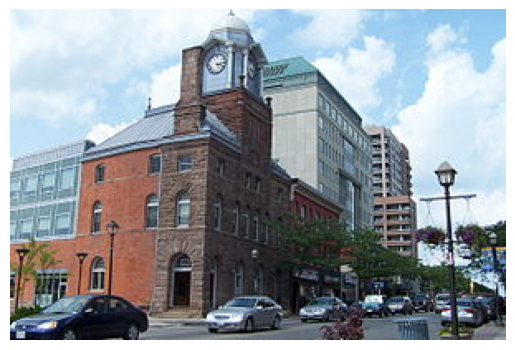}\\ 
\cline{2-5}
 & STS-129 & \vspace{3pt}\includegraphics[width=0.1\textwidth]{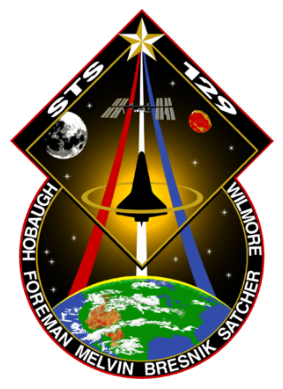} & STS-129 & \includegraphics[width=0.1\textwidth]{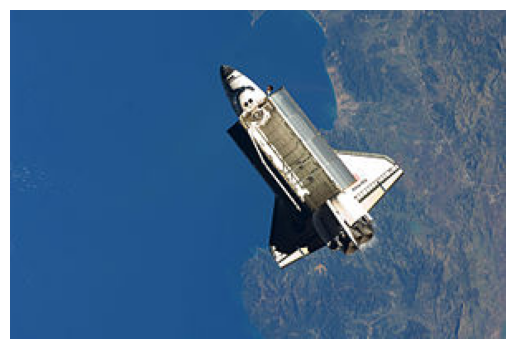}\\
\cline{2-5}
 & Queens of the Stone Age & \vspace{2pt}\includegraphics[width=0.1\textwidth]{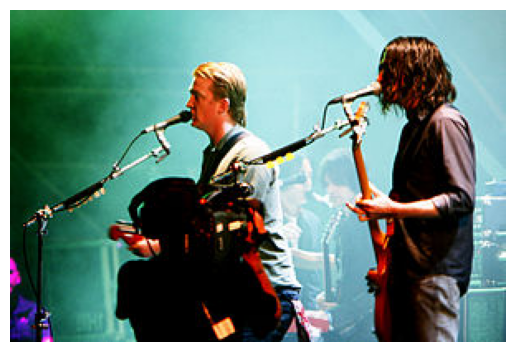} & Queens of the Stone Age & \includegraphics[width=0.1\textwidth]{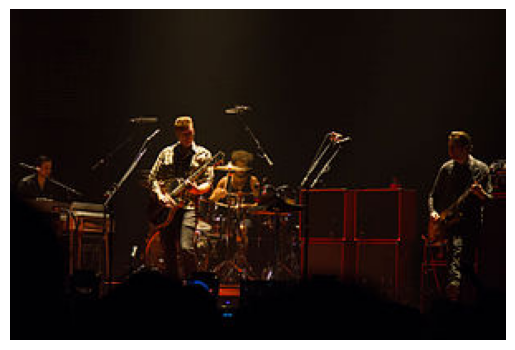}\\
\hline
\end{tabular}}
\label{tab:case study}
\end{table}
To illustrate differences in features frozen at different stages, in Table~\ref{tab:case study}, we select frozen image information during training in the DBP15K\(_{\rm FR-EN}\) dataset arranged in a time series. Early frozen image pairs show significant differences. For example, the "President\_of\_Mexico" entity corresponds to a presidential portrait in the DBP\(_{\rm FR}\) but to a national flag image in the DBP\(_{\rm EN}\). Images will not benefit the process of alignment in this case.
After several epochs of training, the differences between frozen image pairs frozen later are smaller. For example, the "Juliana\_of\_the\_Netherlands" entity, where the image in DBP\(_{\rm FR}\) is a photo of the queen in her youth, while in DBP\(_{\rm EN}\), it is a photo of the queen in her old age. These cases reflect the process of effectively detecting alignment-irrelevant features based on the principle of progressing from easy to difficult cases.

\section{Conclusion}

In this study, we presented the Progressive Modality Freezing (PMF) model to advance Multi-Modal Entity Alignment.
By measuring and evaluating the relevance of various modalities, PMF progressively freezes features deemed less critical, thereby facilitating the integration and consistency of multi-modal features.
Furthermore, we introduced a unified training objective tailored to foster a harmonious contrast between KGs and modalities. 
Empirical evaluations on nine sub-datasets affirm the superiority of PMF and validate the rationale behind the selective freezing of modalities.

\section*{Limitations}
PMF marks a significant advancement in MMEA, yet it encounters limitations in two main aspects.

First, the method for measuring the alignment relevance of modal features by evaluating their usefulness in identifying similar traits across KGs sometimes shows inconsistent reliability. As shown in Figure~\ref{fig:hyper-analy}, a minor decline in H@1 when \(\delta\) is below 0.5 suggesting that essential modal information may be prematurely frozen. Future work should aim to refine the approach for scoring alignment relevance, ensuring a more consistent and accurate identification of pertinent features.

Second, our model predominantly exhibits efficiency in non-iterative learning settings, with its performance often unstable under iterative learning frameworks. This limitation stems from the model's initial design, which is inherently optimized for non-iterative settings, limiting the effectiveness of incorporating iterative strategies like pseudo seeds for enhancing progressive freezing. 
Further attempts will 
explore improved ways to integrate iterative learning with progressive freezing for enhanced results.

\section*{Acknowledgments}
This work was supported by the National Science and Technology Major Project under Grant 2022ZD0120202, in part by the National Natural Science Foundation of China (No. U23B2056), in part by the Fundamental Research Funds for the Central Universities, in part by the State Key Laboratory of Complex \& Critical Software Environment, and in part by the Shanxi Province Special Support for Science and Technology Cooperation and Exchange (202204041101020).

\bibliography{acl2023}

\begin{thebibliography}{25}
\expandafter\ifx\csname natexlab\endcsname\relax\def\natexlab#1{#1}\fi

\bibitem[{Bordes et~al.(2013)Bordes, Usunier, Garcia-Duran, Weston, and Yakhnenko}]{bordes2013translating}
Antoine Bordes, Nicolas Usunier, Alberto Garcia-Duran, Jason Weston, and Oksana Yakhnenko. 2013.
\newblock Translating embeddings for modeling multi-relational data.
\newblock \emph{Advances in neural information processing systems}, 26.

\bibitem[{Chen et~al.(2020)Chen, Li, Wang, Xu, Wang, and Chen}]{chen2020mmea}
Liyi Chen, Zhi Li, Yijun Wang, Tong Xu, Zhefeng Wang, and Enhong Chen. 2020.
\newblock Mmea: entity alignment for multi-modal knowledge graph.
\newblock In \emph{Knowledge Science, Engineering and Management: 13th International Conference, KSEM 2020, Hangzhou, China, August 28--30, 2020, Proceedings, Part I 13}, pages 134--147. Springer.

\bibitem[{Chen et~al.(2022)Chen, Li, Xu, Wu, Wang, Yuan, and Chen}]{chen2022multi}
Liyi Chen, Zhi Li, Tong Xu, Han Wu, Zhefeng Wang, Nicholas~Jing Yuan, and Enhong Chen. 2022.
\newblock Multi-modal siamese network for entity alignment.
\newblock In \emph{Proceedings of the 28th ACM SIGKDD conference on knowledge discovery and data mining}, pages 118--126.

\bibitem[{Chen et~al.(2023{\natexlab{a}})Chen, Chen, Zhang, Guo, Fang, Huang, Zhang, Geng, Pan, Song et~al.}]{chen2023meaformer}
Zhuo Chen, Jiaoyan Chen, Wen Zhang, Lingbing Guo, Yin Fang, Yufeng Huang, Yichi Zhang, Yuxia Geng, Jeff~Z Pan, Wenting Song, et~al. 2023{\natexlab{a}}.
\newblock Meaformer: Multi-modal entity alignment transformer for meta modality hybrid.
\newblock In \emph{Proceedings of the 31st ACM International Conference on Multimedia}, pages 3317--3327.

\bibitem[{Chen et~al.(2023{\natexlab{b}})Chen, Guo, Fang, Zhang, Chen, Pan, Li, Chen, and Zhang}]{chen2023rethinking}
Zhuo Chen, Lingbing Guo, Yin Fang, Yichi Zhang, Jiaoyan Chen, Jeff~Z Pan, Yangning Li, Huajun Chen, and Wen Zhang. 2023{\natexlab{b}}.
\newblock Rethinking uncertainly missing and ambiguous visual modality in multi-modal entity alignment.
\newblock In \emph{International Semantic Web Conference}, pages 121--139. Springer.

\bibitem[{Dietz et~al.(2018)Dietz, Kotov, and Meij}]{dietz2018utilizing}
Laura Dietz, Alexander Kotov, and Edgar Meij. 2018.
\newblock Utilizing knowledge graphs for text-centric information retrieval.
\newblock In \emph{The 41st international ACM SIGIR conference on research \& development in information retrieval}, pages 1387--1390.

\bibitem[{Fang et~al.(2022)Fang, Zhang, Hu, Wu, and Xu}]{fang2022contrastive}
Quan Fang, Xiaowei Zhang, Jun Hu, Xian Wu, and Changsheng Xu. 2022.
\newblock Contrastive multi-modal knowledge graph representation learning.
\newblock \emph{IEEE Transactions on Knowledge and Data Engineering}.

\bibitem[{He et~al.(2016)He, Zhang, Ren, and Sun}]{he2016deep}
Kaiming He, Xiangyu Zhang, Shaoqing Ren, and Jian Sun. 2016.
\newblock Deep residual learning for image recognition.
\newblock In \emph{Proceedings of the IEEE conference on computer vision and pattern recognition}, pages 770--778.

\bibitem[{Li et~al.(2023)Li, Guo, Luo, Ji, Wang, Sheng, and Li}]{li2023attribute}
Qian Li, Shu Guo, Yangyifei Luo, Cheng Ji, Lihong Wang, Jiawei Sheng, and Jianxin Li. 2023.
\newblock Attribute-consistent knowledge graph representation learning for multi-modal entity alignment.
\newblock \emph{arXiv preprint arXiv:2304.01563}.

\bibitem[{Lin et~al.(2022)Lin, Zhang, Wang, Shi, Wu, and Zheng}]{lin2022multi}
Zhenxi Lin, Ziheng Zhang, Meng Wang, Yinghui Shi, Xian Wu, and Yefeng Zheng. 2022.
\newblock Multi-modal contrastive representation learning for entity alignment.
\newblock \emph{arXiv preprint arXiv:2209.00891}.

\bibitem[{Liu et~al.(2021)Liu, Chen, Roth, and Collier}]{liu2021visual}
Fangyu Liu, Muhao Chen, Dan Roth, and Nigel Collier. 2021.
\newblock Visual pivoting for (unsupervised) entity alignment.
\newblock In \emph{Proceedings of the AAAI conference on artificial intelligence}, volume~35, pages 4257--4266.

\bibitem[{Liu et~al.(2019)Liu, Li, Garcia-Duran, Niepert, Onoro-Rubio, and Rosenblum}]{liu2019mmkg}
Ye~Liu, Hui Li, Alberto Garcia-Duran, Mathias Niepert, Daniel Onoro-Rubio, and David~S Rosenblum. 2019.
\newblock Mmkg: multi-modal knowledge graphs.
\newblock In \emph{The Semantic Web: 16th International Conference, ESWC 2019, Portoro{\v{z}}, Slovenia, June 2--6, 2019, Proceedings 16}, pages 459--474. Springer.

\bibitem[{Mao et~al.(2021)Mao, Wang, Wu, and Lan}]{mao2021alignment}
Xin Mao, Wenting Wang, Yuanbin Wu, and Man Lan. 2021.
\newblock From alignment to assignment: Frustratingly simple unsupervised entity alignment.
\newblock \emph{arXiv preprint arXiv:2109.02363}.

\bibitem[{Mousselly-Sergieh et~al.(2018)Mousselly-Sergieh, Botschen, Gurevych, and Roth}]{mousselly2018multimodal}
Hatem Mousselly-Sergieh, Teresa Botschen, Iryna Gurevych, and Stefan Roth. 2018.
\newblock A multimodal translation-based approach for knowledge graph representation learning.
\newblock In \emph{Proceedings of the Seventh Joint Conference on Lexical and Computational Semantics}, pages 225--234.

\bibitem[{Ni et~al.(2023)Ni, Xu, Jiang, Cao, Cao, and Huang}]{ni2023psnea}
Wenxin Ni, Qianqian Xu, Yangbangyan Jiang, Zongsheng Cao, Xiaochun Cao, and Qingming Huang. 2023.
\newblock Psnea: Pseudo-siamese network for entity alignment between multi-modal knowledge graphs.
\newblock In \emph{Proceedings of the 31st ACM International Conference on Multimedia}, pages 3489--3497.

\bibitem[{Pezeshkpour et~al.(2018)Pezeshkpour, Chen, and Singh}]{pezeshkpour2018embedding}
Pouya Pezeshkpour, Liyan Chen, and Sameer Singh. 2018.
\newblock Embedding multimodal relational data for knowledge base completion.
\newblock \emph{arXiv preprint arXiv:1809.01341}.

\bibitem[{Radford et~al.(2021)Radford, Kim, Hallacy, Ramesh, Goh, Agarwal, Sastry, Askell, Mishkin, Clark et~al.}]{radford2021learning}
Alec Radford, Jong~Wook Kim, Chris Hallacy, Aditya Ramesh, Gabriel Goh, Sandhini Agarwal, Girish Sastry, Amanda Askell, Pamela Mishkin, Jack Clark, et~al. 2021.
\newblock Learning transferable visual models from natural language supervision.
\newblock In \emph{International conference on machine learning}, pages 8748--8763. PMLR.

\bibitem[{Simonyan and Zisserman(2014)}]{simonyan2014very}
Karen Simonyan and Andrew Zisserman. 2014.
\newblock Very deep convolutional networks for large-scale image recognition.
\newblock \emph{arXiv preprint arXiv:1409.1556}.

\bibitem[{Sun et~al.(2020{\natexlab{a}})Sun, Cao, Zhao, Wan, Zhou, Zhang, Wang, and Zheng}]{sun2020multi}
Rui Sun, Xuezhi Cao, Yan Zhao, Junchen Wan, Kun Zhou, Fuzheng Zhang, Zhongyuan Wang, and Kai Zheng. 2020{\natexlab{a}}.
\newblock Multi-modal knowledge graphs for recommender systems.
\newblock In \emph{Proceedings of the 29th ACM international conference on information \& knowledge management}, pages 1405--1414.

\bibitem[{Sun et~al.(2017)Sun, Hu, and Li}]{sun2017cross}
Zequn Sun, Wei Hu, and Chengkai Li. 2017.
\newblock Cross-lingual entity alignment via joint attribute-preserving embedding.
\newblock In \emph{The Semantic Web--ISWC 2017: 16th International Semantic Web Conference, Vienna, Austria, October 21--25, 2017, Proceedings, Part I 16}, pages 628--644. Springer.

\bibitem[{Sun et~al.(2020{\natexlab{b}})Sun, Zhang, Hu, Wang, Chen, Akrami, and Li}]{sun2020benchmarking}
Zequn Sun, Qingheng Zhang, Wei Hu, Chengming Wang, Muhao Chen, Farahnaz Akrami, and Chengkai Li. 2020{\natexlab{b}}.
\newblock A benchmarking study of embedding-based entity alignment for knowledge graphs.
\newblock \emph{arXiv preprint arXiv:2003.07743}.

\bibitem[{Xie et~al.(2016)Xie, Liu, Luan, and Sun}]{xie2016image}
Ruobing Xie, Zhiyuan Liu, Huanbo Luan, and Maosong Sun. 2016.
\newblock Image-embodied knowledge representation learning.
\newblock \emph{arXiv preprint arXiv:1609.07028}.

\bibitem[{Xu et~al.(2023)Xu, Xu, and Su}]{xu2023cross}
Baogui Xu, Chengjin Xu, and Bing Su. 2023.
\newblock Cross-modal graph attention network for entity alignment.
\newblock In \emph{Proceedings of the 31st ACM International Conference on Multimedia}, pages 3715--3723.

\bibitem[{Yang(2020)}]{yang2020biomedical}
Zuoxi Yang. 2020.
\newblock Biomedical information retrieval incorporating knowledge graph for explainable precision medicine.
\newblock In \emph{Proceedings of the 43rd International ACM SIGIR Conference on Research and Development in Information Retrieval}, pages 2486--2486.

\bibitem[{Zhu et~al.(2015)Zhu, Zhang, R{\'e}, and Fei-Fei}]{zhu2015building}
Yuke Zhu, Ce~Zhang, Christopher R{\'e}, and Li~Fei-Fei. 2015.
\newblock Building a large-scale multimodal knowledge base system for answering visual queries.
\newblock \emph{arXiv preprint arXiv:1507.05670}.

\end{thebibliography}
\bibliographystyle{acl_natbib}

\appendix

\section*{Appendix}

\section{Detailed setup}\label{sec:detailed setup}

\subsection{Encoders of each modality}\label{app:Encoder Details}
We incorporate four modalities, including structure, image, relationship, and attribute, to implement and evaluate the proposed method. The settings of encoders for each modality are as follows.

The modality of structures describes the topological structure of entities in the knowledge graph. Specifically, the input of structural modal is the connections between various entities in the knowledge graph. The connections can be described by the adjacency matrix of the knowledge graph. Let the adjacency matrix of $\mathcal{G}$ be $A$, we use graph attention networks (GAT) to encode structural information. Therefore, we define:
\begin{equation}
         {h_i}^{str} = \textbf{GAT}(A,\mathbf{e}_i^{str}),
\end{equation}
Where $\mathbf{e}_i^{str}$ is the randomly initialized representation of the entity $e_i$.

The modality of relations consists of the relationships related to each entity in text form. The relational feature of entity $e_i$ is the set of relation names from all the triples in which $e_i$ occurs. Since the number of relations is limited, we use the bag-of-words method to process input features. $\mathbf{e}_i^{rel}$ is defined as a one-hot vector, and the dimension is determined by the number of relations. $\rm{ENC}_{rel}$ consists of a fully connected layer.

Similarly, textual entity attributes constitute an independent modality. Since the number of attributes is also limited, we adopt a processing method similar to the modality of relations. Specifically, $\rm{ENC}_{att}$ is a fully connected layer, and the input $\mathbf{e}_i^{att}$ is defined as a one-hot vector, and the dimension depends on the number of attributes.

The image modality is composed of the visual entity attributes. Encoding images is usually difficult and requires a large amount of images for pre-training. Therefore, we first use a fixed pre-trained vision model~\cite{simonyan2014very, he2016deep, radford2021learning}, to convert the original pixel information into a vector representation $\mathbf{e}_i^{img}$. Then, a trainable fully connected layer is used as $\rm{ENC}_{img}$.

The encoder of relations, attributes, and images can be expressed by:
\begin{equation}
         {h_i}^{m} = \mathbf{FC_m}(\mathbf{e}_i^{m}),
\end{equation}
where $\mathbf{FC_m}$ is a trainable fully connected layer, $m\in\{rel, attr, img\}$.

\subsection{Dataset Statistics}
\label{Dataset Statistics}
Statistics for DBP15K, Multi-OpenEA, and MMKG are presented in Table~\ref{tab:dataset-summary}. The column named "EA pairs" indicates pre-aligned entity pairs. Note that not every entity is paired with images or counterparts in the alternate KG. 
\begin{table*}[!ht]
\centering
\small
\caption{Statistical details for MMEA datasets.}
\label{tab:dataset-summary}
\begin{tabular}{
  c|
  c|
  c c c c c c c
}
\toprule
{Dataset} & {KG} & {\# Ent.} & {\# Rel.} & {\# Attr.} & {\# Rel. Triples} & {\# Attr. Triples} & {\# Image} & {\# EA pairs} \\
\midrule
\multirow{2}{*}{DBP15K\(_{ZH-EN}\)} & ZH (Chinese) & 19388 & 1701 & 8111 & 70414 & 248035 & 15912 & \multirow{2}{*}{15000}
\\
 & EN (English) & 19572 & 1323 & 7173 & 95142 & 343218 & 14125 &
\\
\hline
\multirow{2}{*}{DBP15K\(_{JA-EN}\)} & JA (Japanese) & 19814 & 1299 & 5882 & 77214 & 248991 & 12739 & \multirow{2}{*}{15000}
\\
& EN (English) & 19780 & 1153 & 6066 & 93484 & 320616 & 13741 & 
\\
\hline
\multirow{2}{*}{DBP15K\(_{FR-EN}\)} & FR (French) & 19661 & 903 & 4547 & 105998 & 273825 & 14174 & \multirow{2}{*}{15000}
\\
& EN (English) & 19993 & 1208 & 6422 & 115722 & 351094 & 13858 & 
\\
\hline
\multirow{2}{*}{Multi-OpenEA\(_{EN-FR}\)}& EN (English) & 15000 & 267 & 308 & 47334 & 73121 & 15000 & \multirow{2}{*}{15000}
\\
& FR (French) & 15000 & 210 & 404 & 40864 & 67167 & 15000 & 
\\
\hline
\multirow{2}{*}{Multi-OpenEA\(_{EN-DE}\)}& EN (English) & 15000 & 215 & 286 & 47676 & 83755 & 15000 & \multirow{2}{*}{15000}
\\
& DE (German) & 15000 & 131 & 194 & 50419 & 156150 & 15000 & 
\\
\hline
\multirow{2}{*}{Multi-OpenEA\(_{DW-V1}\)}& DBpedia & 15000 & 248 & 342 & 38265 & 68258 & 15000 & \multirow{2}{*}{15000}
\\
& Wikidata & 15000 & 169 & 649 & 42746 & 138246 & 15000 & 
\\
\hline
\multirow{2}{*}{Multi-OpenEA\(_{DW-V2}\)}& DBpedia & 15000 & 167 & 175 & 73983 & 66813 & 15000 & \multirow{2}{*}{15000}
\\
& Wikidata & 15000 & 121 & 457 & 83365 & 175686 & 15000 & 
\\
\hline
\multirow{2}{*}{FBDB15K} & FB15K & 14951 & 1345 & 116 & 592213 & 29395 & 13444 & \multirow{2}{*}{12846}
\\
& DB15K & 12842 & 279 & 225 & 89197 & 48080 & 12837 & 
\\
\hline
\multirow{2}{*}{FBYG15K} & FB15K & 14951 & 1345 & 116 & 592213 & 29395 & 13444 & \multirow{2}{*}{11199}
\\
& YAGO15K & 15404 & 32 & 7 & 122886 & 23532 & 11194 & 
\\
\bottomrule
\end{tabular}
\end{table*}

\subsection{Metric Details}
We evaluate the performance of entity alignment tasks using two common metrics: Hits@n and MRR.

\textbf{Hits@n} is a widely used performance metric in entity alignment tasks. It is calculated as follows: Given a total number of entity pairs \(N\), for each entity in graph \(\mathcal{G}\), compute its similarity with potential matching entities in graph \(\mathcal{G'}\) and rank them according to their scores. If the correct entity's rank is within the top \(n\) positions, the Hits@n count is incremented. The value of Hits@n is the percentage of the Hits@n count out of the total number of entity pairs \(N\). Higher Hits@n values indicate better alignment performance.

\begin{equation}
    Hits@n = \frac{1}{N} \sum_{i=1}^{N} I(rank_i \leq n)
\end{equation}

\textbf{MRR} (Mean Reciprocal Rank) is a general evaluation metric for search algorithms and, when applied to entity alignment tasks, calculates the similarity between each entity in graph \(\mathcal{G}\) and potential matching entities in graph \(\mathcal{G}\), ranking them according to their scores. If the correct entity is ranked \(n\), the score is \(\frac{1}{n}\). The sum of scores for all entities yields the MRR score. MRR reflects the overall performance of the entity alignment algorithm, with higher MRR scores indicating better alignment results.
\begin{equation}
    MRR = \frac{1}{N} \sum_{i=1}^{N} \frac{1}{rank_i}
\end{equation}

\subsection{Implementation Details}
For a fair comparison with recent works~\cite{chen2023meaformer,chen2023rethinking}, we have standardized our experimental setup as follows:
\begin{itemize}
\item 
\textbf{Iterative Learning}: We adopt the approach of~\cite{lin2022multi} and employ a probation method for iterative training. This method collects pairs of mutual nearest entities every 5 epochs. Pairs that consistently remain nearest neighbors for 10 rounds are added to the training set.
\item
\textbf{Multi-modal Encoders}: 
For relations and attributes, we use the Bag-of-Words (BoW) model to encode relations (\(x_r\)) and attributes (\(x_a\)) as fixed-length vectors with a dimension of \(d_r = 1000\). For entity names' surface information, we use 300-dimensional GloVe vectors and character bigrams. We enhance this method by applying machine translations to entity names, following~\cite{mao2021alignment}.
For images, we use ResNet-152~\cite{he2016deep} as pre-trained visual encoders with a vision feature dimension (\(d_v\)) of 2048 in DBP15K. For Multi-OpenEA, CLIP~\cite{radford2021learning} is used with \(d_v = 512\). In FBDB15K/FBYG15K, VGG-16~\cite{simonyan2014very} is employed with \(d_v = 4096\). It should be noted that we assign a zero vector as the embedding for missing images to minimize their impact on cross-modal learning.
\item 
\textbf{Hyper Parameters}:
We unify the hidden layer dimensions across all networks to 300. 
The parameter \(\tau\) is set to 0.05 to emphasize challenging negative samples in the contrast loss. The dynamic threshold \(\delta\) in Eq.~\ref{exp:confidence} for progressively freezing entities begins at 0.1 and increases by 1.2 until it reaches a maximum of 0.9. For the structure, relation, attribute, and image modalities, \(\beta_m\) values in Eq.~\ref{cross-modal loss} are respectively set to 0.1, 0.1, 0.1, and 10, scaling the learning rate of modality features.
\item
\textbf{Computational Overhead}:
Our experiments, conducted on a Tesla V100 SXM2 32GB GPU, demonstrate the superior efficiency of our model, which completes training in only 20 minutes. This is notably faster than the baseline model, MEAformer~\cite{chen2023meaformer}, at 33 minutes, and UMAEA~\cite{chen2023rethinking} at 29 minutes. The number of learnable parameters is 13.1M for DBP15K and Multi-OpenEA, and 9.9M for MMKG.
\end{itemize}

\section{Detailed analysis}\label{app:detailed analysis}
\subsection{Low Resource }
\begin{figure}[h]
    \centering
    \includegraphics[width=1\linewidth]{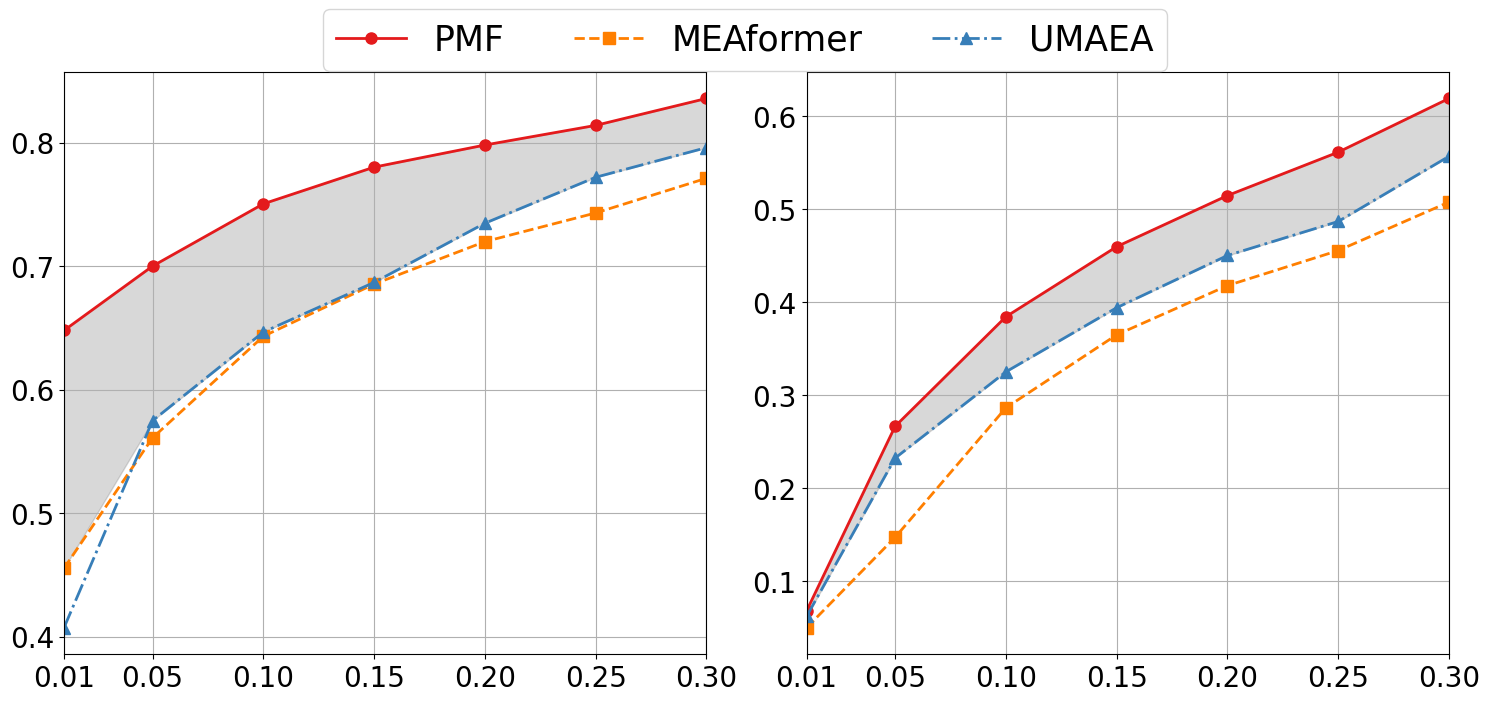}
    \caption{H@1 performance with different ratios of seed alignments ranging from 0.01 to 0.3 on DPB15K\(_{ZH-EN}\) (left) and FBDB15K (right). The x-axis denotes ratios and the y-axis denotes H@1.
}
    \label{fig:low resource}
\end{figure}
To assess the stability of the proposed method under conditions of limited seed alignments, we conducted evaluations on two distinct datasets--DBP15K\(_{ZH-EN}\) and FBDB15K, using seed alignment ratios varying from 0.01 to 0.30. Figure~\ref{fig:low resource} illustrates a clear gap between performances as the ratio escalates. Notably, with merely 1\% seed alignments in the DBP15K\(_{ZH-EN}\) dataset, our model was able to achieve a H@1 score of .648, markedly outperforming the MEAformer, which scored .456. This result highlights the model's robustness and its considerable promise for applications in few-shot entity alignment.

\subsection{Analysis of Freezing Process}
\begin{figure}[h]
    \centering
    \includegraphics[width=1\linewidth]{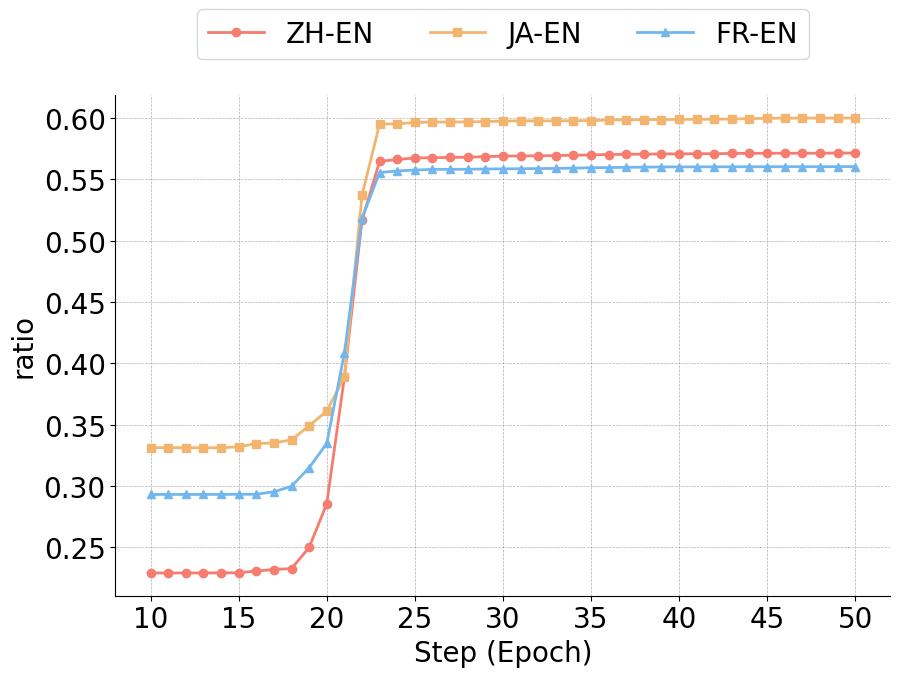}
    \caption{The ratio of entities with images that are frozen during the training epochs in DBP15K.}
    \label{fig:frozen ratio}
\end{figure}
To delve deeper into the impacts of the progressive manner, we take the freezing process of image modality on the DBP15K dataset as an example, analyzing the trend of the proportion of entities being frozen as training epochs change. Figure~\ref{fig:frozen ratio} indicates that as training epochs increase, the proportion of entities frozen shows an exponential rising trend, ultimately stabilizing at about 55\%. The high rate of frozen images is partly due to a high image missing rate of up to 30\% in the DBP15K dataset, and it also reflects high ambiguity in image information within multi-modal knowledge graphs.
\end{document}